\documentclass[runningheads]{llncs}

\usepackage[T1]{fontenc}
\usepackage{multirow}
\usepackage{graphicx}
\usepackage{booktabs}
\usepackage{adjustbox}
\usepackage{array}
\usepackage{amsmath}
\usepackage{graphicx}
\usepackage[table,xcdraw]{xcolor}
\usepackage{pifont}
\usepackage{float}
\usepackage{amsfonts}
\begin{document}
\title{Interaction-Guided Two-Branch Image Dehazing Network}
%
%
\author{Huichun Liu\inst{1} \and
Xiaosong Li\inst{1, }\thanks{Corresponding Author} \and
Tianshu Tan\inst{2}}
\authorrunning{H.Liu et al.}
%
\institute{School of Physics and Optoelectronic Engineering, Foshan University, Foshan 528225, China \and
Hong Kong University of Science and Technology, Hong Kong, China\\
\email{Feecuin@outlook.com, lixiaosong@buaa.edu.cn, ttanad@connect.ust.hk} \\
}
\maketitle              
\begin{abstract}
Image dehazing aims to restore clean images from hazy ones. Convolutional Neural Networks (CNNs) and Transformers have demonstrated exceptional performance in local and global feature extraction, respectively, and currently represent the two mainstream frameworks in image dehazing. \textbf{In this paper, we propose a novel dual-branch image dehazing framework that guides CNN and Transformer components interactively.} We reconsider the complementary characteristics of CNNs and Transformers by leveraging the differential relationships between global and local features for interactive guidance. This approach enables the capture of local feature positions through global attention maps, allowing the CNN to focus solely on feature information at effective positions. The single-branch Transformer design ensures the network's global information recovery capability. Extensive experiments demonstrate that our proposed method yields competitive qualitative and quantitative evaluation performance on both synthetic and real public datasets. Codes are available at https://github.com/Feecuin/Two-Branch-Dehazing

\end{abstract}
\section{Introduction}
\label{sec:intro}
Haze is caused by small particles in the atmosphere scattering light, which reduces the visibility of objects and leads to a decline in the performance of visual systems in practical tasks such as autonomous driving, object detection, and drone aerial photography. Image dehazing technology\cite{T1,T2,T3,T4} can eliminate the influence of haze, restore scene visibility, and provide high-quality images to visual systems.
In early image dehazing techniques, the relationship between hazy  and clean images was typically described by the following model\cite{M1}:
\begin{equation}
I=J(\hat{x})t(\hat{x})+A(1-t(\hat{x})),
\label{eq1}
\end{equation}
where \(\hat{x}\) is the 2D spatial location, \(I\) is the captured hazy image, \(J\) is the clean image, \(A\) is the global atmospheric light, and \(t(\hat{x})\) is the transmission map, which is expressed as
\begin{equation}
t(x)=e^{-\beta d(\hat{x})},
\label{eq2}
\end{equation}
the transmission map \(t(\hat{x})\) depends on the depth of the scene \(d(\hat{x})\) and the haze density coefficient \(\beta\). According to this model formula, many priors\cite{P1,P2,P3} were proposed in the early stages to constrain the ill-posedness it brings. However, such prior-based approaches rely on empirical knowledge and are difficult to adapt to different scenarios, and may produce artifacts in areas where priors are not satisfied.
With the rise of deep-learning-based dehazing methods, many CNN-based dehazing algorithms have emerged\cite{c6,c2,c1}, which can achieve better performance than prior-based methods. However, the convolutional mechanisms of CNNs determines that they are limited by smaller receptive fields. \\
\indent A pure convolutional model causes a network to focus excessively on the local features of an image (e.g., edges and texture information), which is not conducive to the overall restoration of the image. Recently developed Transformer models\cite{TS2} have achieved superior global feature extraction capabilities compared to CNN on various computer vision tasks. Its attention mechanism ensures that it has good global feature extraction capabilities. However, Transformers often lead to unwanted blurring and rough details during image reconstruction. Existing method\cite{irt1} do not consider the feature correlation between CNN and Transformer, resulting in feature redundancy. \textbf{To combine the advantages of CNN and Transformer for feature extraction}, we propose an interactive guidance method that utilizes the ability of a Transformer to extract global features to provide accurate global information and guide a CNN to focus on detailed information within an effective feature space.\\
\indent CNN are able to identify most of the useful details of images, and considering that Transformer will intersect with CNN during feature extraction, adding features directly will lead to information redundancy. Therefore, the down-sampling operation is introduced into the Transformer branch to distinguish the features extracted by the two branches as much as possible, avoid redundancy caused by repeated extraction, and improve the performance of the model. The details lost by downsampling will be compensated in the CNN branch. The proposed method makes full use of the complementary advantages of CNN and transformer to provide high-quality dehazing results with limited computing resources. In summary, our contributions are as follows:
\begin{itemize}
    \item We propose an interaction-guided dual-branch image dehazing framework that utilizes the global information provided by the Transformer to guide the CNN to focus on local details effectively, while the Transformer branch ensures the ability of the network to recover global information.
\end{itemize}
\begin{itemize}
    \item Our method effectively reduces the redundant information generated by the repeated feature extraction of Transformer and CNN, thereby improving the performance of the two-branch model.
\end{itemize}
\begin{itemize}
    \item Extensive experiments on existing synthetic and real datasets consistently confirm the superiority of our model, and the ablation experiments demonstrate the effectiveness of each module.
\end{itemize}
\section{Related Works}
\label{sec:format}
\begin{figure*}
\centering
\includegraphics[width=1.0\linewidth]{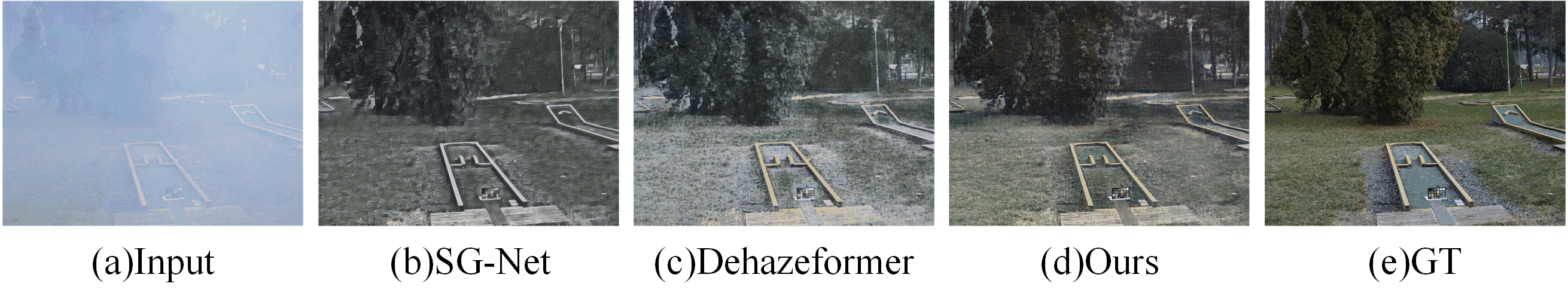}
\caption{(a) hazy image, (b) and (c) represent the  dehazing results of CNN method and the Transformer method, respectively, (d) Dehazing result of our method, and (e) groundtruth image.}
\label{Fig1}
\end{figure*}

\textbf{\textit{Single Image Dehazing.}} Image dehazing has always been a challenging and important task in computer vision and image processing. It can restore hazy images to clean images and has important applications in many fields. In recent years, many image dehazing methods\cite{P1,P2,P3} have been proposed, and early methods generally considered the effects of particle scattering in the atmosphere. They attempted to derive the parameters of atmospheric scattering using mathematical formulas, but these manually derived priors such as dark channel priors\cite{P1}, color attenuation priors\cite{P2}, and non-local priors\cite{P3} were derived based on empirical knowledge and are often difficult to adapt to diverse scenarios. When the scene does not meet these priors, these prior-based dehazing algorithms often output some results that do not meet expectations. For example, the dark channel prior\cite{P1} cannot handle the sky region, which can lead to image distortion. The saturation line prior (SLP)\cite{SLP} reveals the linear relationship between the inverse of the saturation component and the brightness component in the local pixels of the normalized hazy image, and proposes a novel image dehazing framework to exploit the linear distribution of local pixels, which helps to improve the transfer estimation for better detail restoration and color preservation. In recent years, many deep learning-based methods for image dehazing have emerged, and these methods have gradually become mainstream. Early deep learning methods were still related to Atmospheric Scattering Model(ASM), for example, the CNN model DehazeNet\cite{c1} aims to estimate the transmission map t, 
 and then substitute the estimated transmission map into the ASM for calculation to obtain dehazed images. DehazeNet represents a pioneering effort in image dehazing.\\
\indent Afterwards, AOD-Net\cite{c3} simultaneously estimates the transmission map t and atmospheric light A, and obtains  the restored haze-free image through ASM. However, methods based on prior estimation often have some bias. Many recently proposed deep learning models do not require parameter estimation to be substituted into ASM for computation. Instead, these models directly restore blurry and hazy images to clean images. For example, GridDehazeNet\cite{c4} proposed that learning to recover images is better than directly estimating t, and designed attention-based multi-scale estimators to achieve dehazing. FFA-Net\cite{c5} introduced feature attention (FA) blocks, utilizing pixel and channel attention to improve the model's dehazing performance. MSBDN\cite{c6} skillfully combines the enhancement strategy and back-projection technology for image dehazing, and proposes a multi-scale enhanced dehazing network. SG-Net\cite{SG-Net} proposes a novel end-to-end network to restore haze-free images, and the simple and efficient SG mechanism can be embedded into existing network families at will, with only a little extra time consumption while improving accuracy. In the early days, CNN dominated most computer vision tasks. In recent years, Transformer\cite{TS3,TS4} has been widely used in computer vision, such as object detection, image segmentation, and other tasks. \\
\indent Recently, many Transformer-based deep learning methods have emerged for image restoration\cite{irt1,irt2,irt3,325,kungong}. For example, Restormer\cite{irt2} designed multi-head attention and feedforward networks to enable models to capture remote pixel interactions, and DehazeFormer\cite{T3} adds reflection filling to the sliding window mechanism to reduce the loss of edge information in Swin Transformer\cite{TS2} when processing hazy images. Fourmer\cite{fourmer} adds a Fourier-based general image degradation prior to the core structure of Fourier spatial modeling and Fourier channel evolution, which provides new insights into the design of image inpainting based on global modeling. \\
\indent In recent years, some image dehazing algorithms with a two-branch structure have emerged, combining the advantages of CNN and Transformer to improve their dehazing performance. However, none of these two-branch dehazing algorithms considers the relationship between the features in the two-branch setup, and do not design the network in terms of feature differences and feature redundancy caused by repeated extraction. In contrast, we scrutinize the complementary properties between the two, integrate their strengths to enhance feature extraction, and also utilize the different relationships between features for interactive guidance to improve network performance. As shown in Fig.\ref{Fig1}, from left to right are the hazy image, and the dehazing results of  SG-Net\cite{SG-Net}, Dehazeformer\cite{T3} and the proposed method, and the rightmost is the ground truth image. It can be seen that the image details recovered by SG-Net are clear, but the global consistency is poor, the overall tone and brightness are not natural enough, and there is a loss of details. The overall results appears fuzzy; Dehazeformer is superior to SG-Net in terms of global features and hue, and the overall image is harmonious, but its local details processing is weak, and some residual noise remains. Our method combines the advantages of CNN and Transformer model, and the recovered image not only maintains global consistency and naturalness but also handles local details carefully. This results in an overall high-definition image with significantly reduced noise.

\begin{figure*}[ht]
\centering
\includegraphics[width=1.0\linewidth]{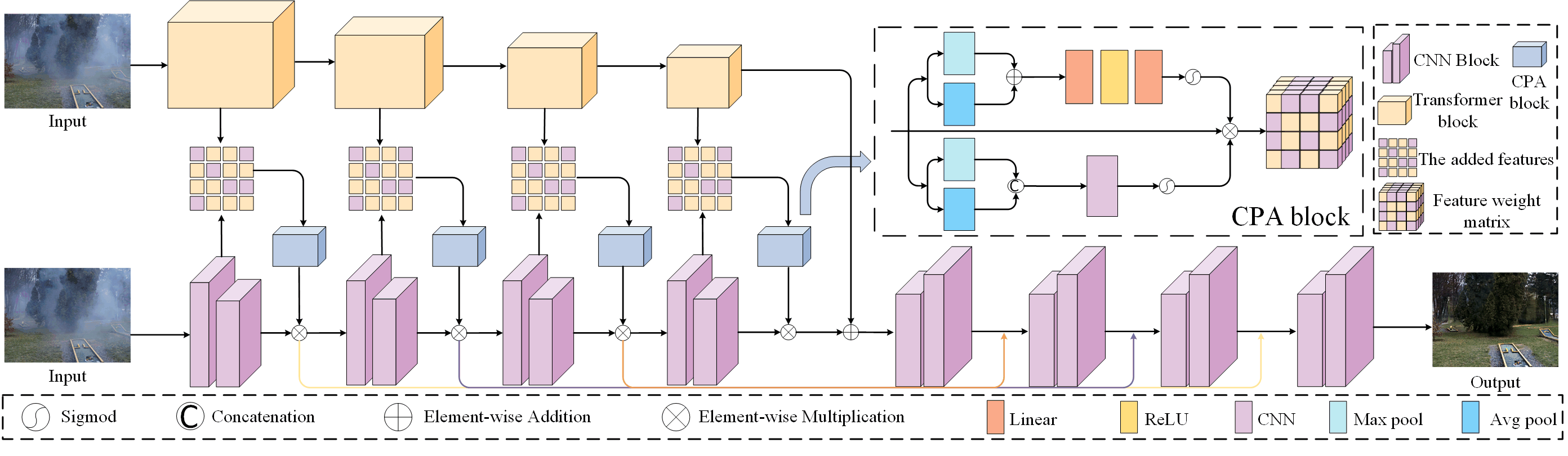}
\caption{Our method utilizes Transformer and CNN branches, where the output features in the middle of each layer are utilized in a CPA generated attention map to guide the CNN. The CNN results are combined with the results of the Transformer branch to perform CNN upsampling to recover image details.}
\label{Fig2}
\end{figure*}

\section{Proposed Method}
\label{sec:pagestyle}
An overview of the proposed framework is presented in Fig.\ref{Fig2}. A blurred image is input into two branches, namely a CNN and a Transformer, to extract local and global features, respectively. The global information extracted by each layer of the Transformer is used to guide the CNN to focus on effective local information. Downsampling is performed on the blurred image before it enters the Transformer block to expand its receptive field and reduce computational complexity. The image details are then restored by the CNN decoder, and a final haze-free image is obtained. Various experiments have demonstrated that our network provides good performance on real datasets.\\
\\
\textbf{Network Structure}\\
\\
\textit{\textbf{Global Perception Module.}} We utilize a DehazeFormer module to extract global features, and its improvements in normalization layer and spatial aggregation scheme make it more efficient in dehazing than the original Swin-Transformer. The normalization layer preserves the mean and standard deviation of the original image, ensuring that the restored image has the same contrast and brightness as those of  image. The normalization layer is represented as follows:
\begin{equation}
y = F\left(\frac{x - \mu}{\sigma}  \gamma + \beta \right) \cdot (\sigma  W_\gamma + B_\gamma) + (\mu  W_\beta + B_\beta ) ,
\label{eq3}
\end{equation}
the feature map  \(x \in \mathbb{R}^{b \times n \times c}\) ,where \(n=h \times w\) (i.e. height and width), \( \mu\) and \(\sigma\) represents the mean and standard deviation, respectively; \(\gamma\) and \( \beta\) denote scaling factors and biases, respectively; \(W_\beta\) and \(W_\gamma\)and \(B_\beta\) and \(B_\gamma\) are the weights and bias used for transformation \(\mu\) and \(\sigma\), respectively.\\
\indent The sliding window mechanism of DehazeFormer utilizes reflection filling to ensure that the size of the edge window is the same as that of the set window to prevent missing edge information in an image, which can improve the performance of the network. Additionally, DehazeFormer adds a layer of convolution after ordinary aggregation because the attention mechanism aggregates information within the edge window while ignoring information between windows. Therefore, convolutional operations can be used to aggregate information between neighboring windows, which is represented as follows:
\begin{equation}
\text{Aggregation}(Q, K, V) = \text{Softmax}\left(\frac{QK^T}{\sqrt{d}} + B \right) + \text{Conv}(\hat{V}),
\label{eq4}
\end{equation}
where \(\hat{V} \in \mathbb{R}^{b \times h \times w \times c}\), is \(V\) prior to window division.\\
\indent In the proposed method, we use four transformer layers to extract global features, with a downsampling layer preceding the transformer input. The receptive field of the Transformer branch is increased, and the redundant information generated by the repeated feature extraction of the Transformer and CNN branches is reduced, thereby improving the model's performance.\\
\textit{\textbf{Local Perception Module.}} To achieve the function of extracting local features, we introduce CNN as another branch of the model. Due to the fact that CNN is mainly based on local perception, its convolutional mechanism also determines that it has a smaller receptive field, which can effectively extract local detail information. However, pure convolutional models can lead to the excessive extraction of image edges and texture details, which can lead to a decrease in model performance and an increase in computational costs. Therefore, we propose leveraging the advantages of the Transformer to extract global features to guide the CNN in local feature extraction. We use a four-layer CNN corresponding to the Transformer structure, where each layer has two outputs. One output is added to the output of the Transformer, and then an attention map is generated by the channel and pixel attention block (CPA). The attention map is then multiplied by the other output of the current CNN layer, and the results are input into the next layer of the CNN to provide global guidance for local extraction, thereby helping the CNN extract detail information more effectively. In the decoder, we utilize four convolution layers, corresponding to the number of layers in the Transformer and CNN branches.\\
\indent Image upsampling is performed by the decoder to restore detail information. Skip connections are made with the CNN in the dual branches to help preserve the details of the original image. These connections also serve to avoid the loss of detail information caused by network training and improve overall network performance. Finally, this process outputs a clean and haze-free image.\\
\textbf{\textit{Channel and pixel attention.}} We extracted the features between each layer of the dual branches and add them together to obtain effective information regarding the entire image space to guide the CNN. After these features pass through the CPA block, a weight matrix is obtained to guide the CNN. The CPA block comprises channel attention\cite{channel} and pixel attention\cite{pixel}, with \(x\) and \(y\) as input and output, respectively. The channel attention formula is defined as follows:
\begin{equation}
\text{channel}_{\text{att}} = \text{Sigmoid}\left( \text{FC}(\text{avgpool}(x), \text{maxpool}(x)) \right),
\label{eq5}
\end{equation}
\(\text{channel}_{\text{att}}\) is the channel attention weight of the feature map. Channel attention models the global information of the entire channel, capturing the global importance of each channel in the input feature map, reducing redundant information, and improving  the attention paid to key features.
The pixel attention formula is defined as follows:
\begin{equation}
\text{pixel}_{\text{att}} = \text{Sigmoid}\left( \text{conv1}(\text{mean}(x), \text{max}(x)) \right),
\label{eq6}
\end{equation}
\(\text{pixel}_{\text{att}}\) is the pixel attention weight of the feature map. Pixel attention models the local information at each pixel location and can capture the importance of each location in the input feature map, enhancing the model's perception of the local features of the input image, with enhanced attention to detail information.
\begin{equation}
y = \text{channel}_{\text{att}} \times \text{pixel}_{\text{att}} \times x ,
\label{eq7}
\end{equation}
The combination of channel attention and pixel attention leverages their respective advantages, such that the model can pay attention to both channels and pixels at the same time, and has the ability to extract both local and global features, which can better guide the CNN to capture information in the effective feature space, thus improving the performance of the model.

\section{Experiment}
\label{sec:typestyle}

In this section, we conducted extensive experiments to demonstrate the effectiveness of our proposed method. Firstly, we introduce the experimental setup, and then compare it with advanced dehazing methods on both synthetic and real datasets. Additionally, ablation studies are presented to demonstrate the effectiveness of each component of the proposed model.
\subsection{Experimental Setups}
\textit{\textbf{Training Detail.}} The proposed method was implemented in Python on an Nvidia GeForce RTX 3090 GPU. We used the Adam optimizer with default parameters to optimize our algorithm, setting the initial learning rate to 0.0001 and using only L1 loss. We use randomly cropped image blocks for training, gradually scaling up the size of the image blocks from 128\(\times\) 128 to the full size during training.\\
\textit{\textbf{Dataset.}} Experiments were conducted on the synthetic dataset RESIDE-6K\cite{6k}, the real dataset NH-HAZE\cite{nh}, and DENSE-HAZE\cite{dense}. RESIDE-6K is a mixed dataset of indoor and outdoor images on RESIDE, hence it is called SOTS mix. Its training set includes 3000 OTS image pairs and 3000 ITS image pairs, and its test set is also divided into mixed indoor and outdoor image pairs, with a total of 1000 image pairs combined. The DENSE-HAZE dataset and NH-HAZE dataset are both composed of 45 training images, five validation images, and five test images. The haze of DENSE-HAZE is dense and uniform, while the haze of NH-HAZE is dense and uneven.\\
\textit{\textbf{Compared Methods and Metrics.}} To demonstrate the effectiveness of our method, we compare it with GridDehazeNet\cite{c4}, FFA-Net\cite{c5}, MSBDN\cite{c6}, PSD\cite{PSD}, SG-Net\cite{SG-Net}, D4\cite{D4}, SLP\cite{SLP}, Dehazeformer\cite{T3} (Dehazeformer-T variant) and fourmer\cite{fourmer} on synthetic and real data. If no pretrained model is provided, we retrain the model using the authors' code. Otherwise, we evaluate them using their online code for a fair comparison. All of these representative methods are selected for visual comparison. For the quantitative evaluation of image quality assessment, we use the commonly used PSNR, SSIM, Entropy, and LPIPS to compare the performance of each method.
\subsection{Experimental Results}
\begin{figure*}
\centering
\includegraphics[width=1.0\linewidth]{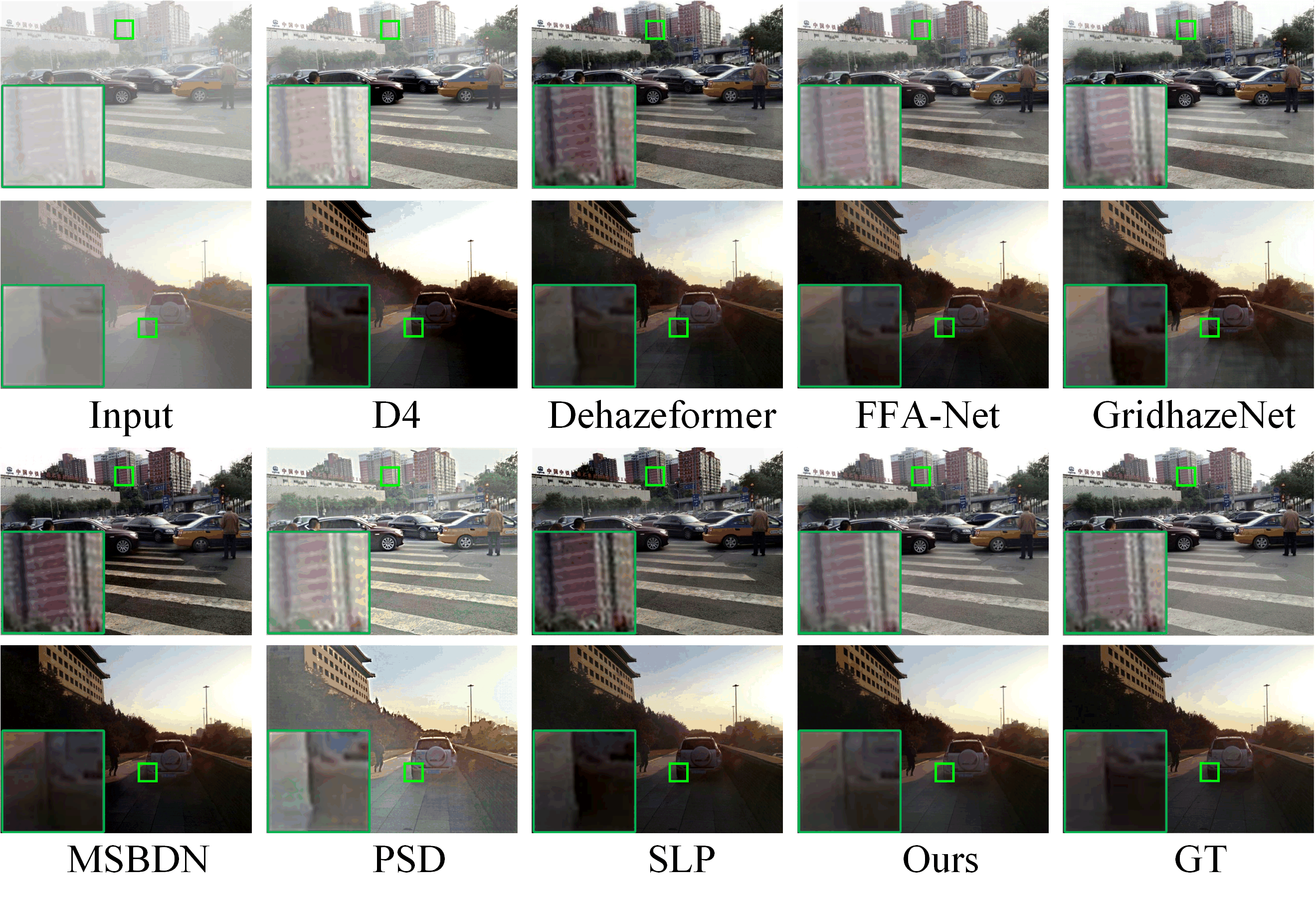}

\caption{Visual comparison of outdoor scenes of different dehazing methods on RESIDE-6K dataset.  (Zooming in can obtain a clearer view)}
\label{fig3}
\end{figure*}

\begin{figure*}
\centering
\includegraphics[width=1.0\linewidth]{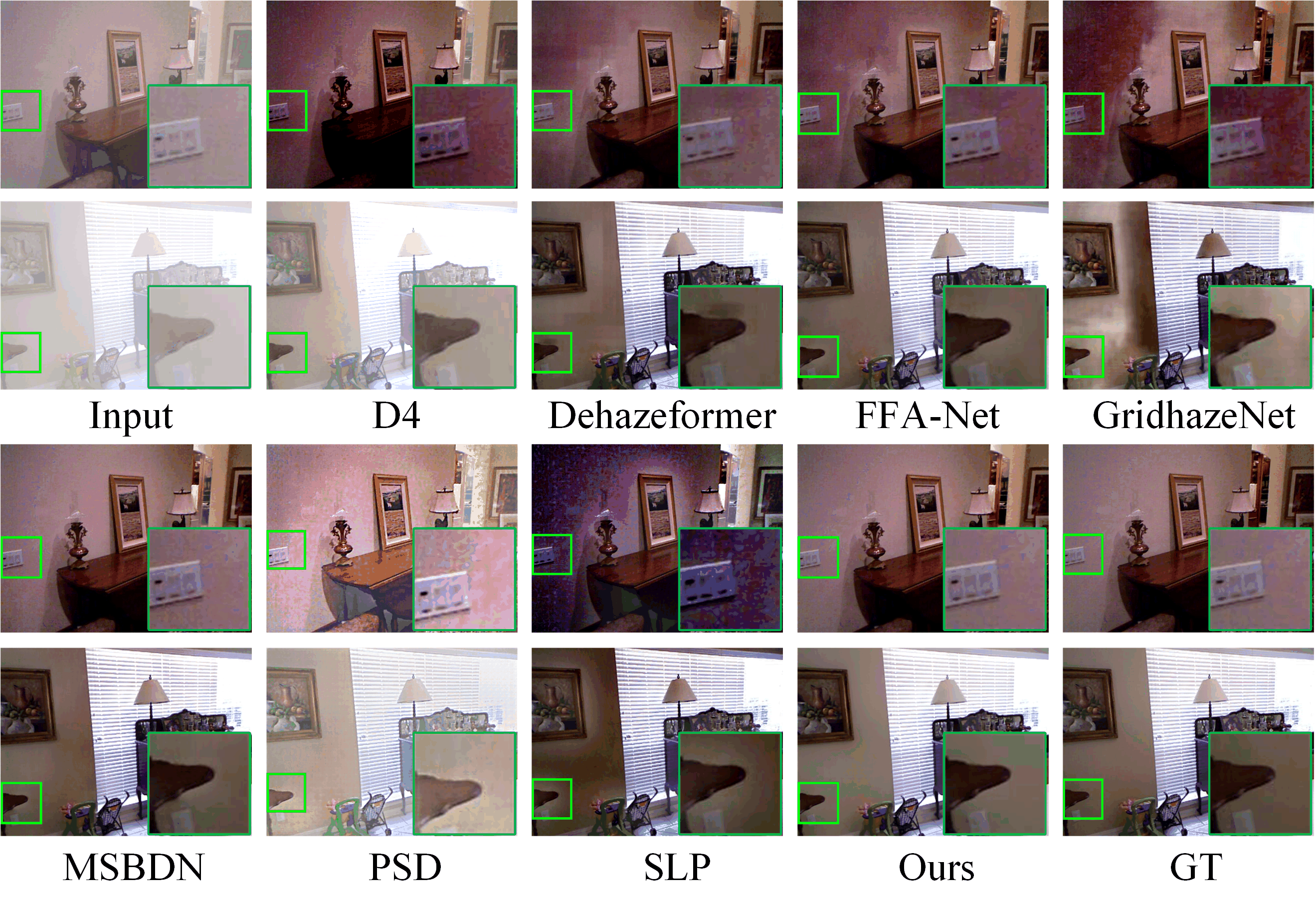}

\caption{Visual comparison of indoor scenes of different dehazing methods on RESIDE-6K dataset. (Zooming in can obtain a clearer view)}
\label{fig4}
\end{figure*}

\textbf{\textit{Results on the Synthetic Datasets.}}
We first test on the synthetic hazy image dataset RESIDE-6K. The images contained in RESIDE-6K can be divided into two types: indoor and outdoor, and the dehazing results of different methods are shown in Fig.\ref{fig3} and \ref{fig4}.

As shown in Fig.\ref{fig3}, there are significant differences between the restored images and the ground truth images for all comparison methods in the outdoor scene. There are residual haze and some undesired details (artifacts, blur, etc.), such as D4,GridDehazeNet and PSD, and most methods have color bias, such as MSBDN and SLP. As shown in Fig.\ref{fig4}, for indoor scenes, almost all images of FFA-Net, SLP, and Dehazeformer have problems such as color deviation (red or purple), blurred details, and low contrast. In contrast, for both outdoor and indoor scenes, the images generated by our method are closest to real haze-free images in terms of details as well as overall image tone.

\begin{table}
\scriptsize
\centering
\caption{Quantitative comparison of the proposed algorithm and different comparison methods on the  RESIDE-6K dataset.  \textbf{Bold} is the best, \textcolor{red}{Red} is the second.}
\label{table1}
\scalebox{1.15}{
\begin{tabular}{c||c||c|c|c|c}
\toprule
\multirow{3}{*}{\textbf{Methods}} & \multirow{3}{*}{\textbf{Venue\&  Year}} & \multicolumn{4}{c}{\textbf{RESIDE-6K}} \\
\cmidrule(lr){3-6}
&& $PSNR\uparrow$ & \ $SSIM\uparrow$  & $Entropy\uparrow$ &  $LPIPS\downarrow$  \\
\midrule
GridDehazeNet \cite{c4} & \textit{ICCV’19} 
&25.65        &0.9371  &   7.4597      & 0.1915  \\

FFA-Net \cite{c5} & \textit{AAAI’20}
&27.26       &0.9567        & 7.4151     &0.1757\\

MSBDN \cite{c6} & \textit{CVPR’20 }
&27.44        &0.9511     &7.4309       &0.1751 \\

PSD \cite{PSD} & \textit{CVPR'21}
&15.47           &0.8149          &\textbf{7.4673}       &0.1697 \\

SG-Net \cite{SG-Net} & \textit{ACCV‘22}
&-       &-        & -            & -     \\

D4 \cite{D4} & \textit{CVPR 22}
&18.97     &0.8422         & 6.5999   &0.1635\\

SLP \cite{SLP} & \textit{TIP'23}
&21.35      &0.9261          &7.3925    &\textcolor{red}{0.0978}     \\

Dehazeformer-T \cite{T3} & \textit{TIP'23}
&\textbf{30.36}         &\textbf{0.9730}      &\textcolor{red}{7.4419}     &\textbf{0.0303}      \\

Fourmer\cite{fourmer} & \textit{ICML'23}
&-   &-         &-            &-            \\

\midrule
\rowcolor{gray!20} \textbf{Ours} & -
& \textcolor{red}{30.20  } & \textcolor{red}{0.9643} & 7.4325 & 0.1749  
\\
\bottomrule
\end{tabular}}
\end{table}

\begin{figure*}
\centering
\includegraphics[width=1.0\linewidth]{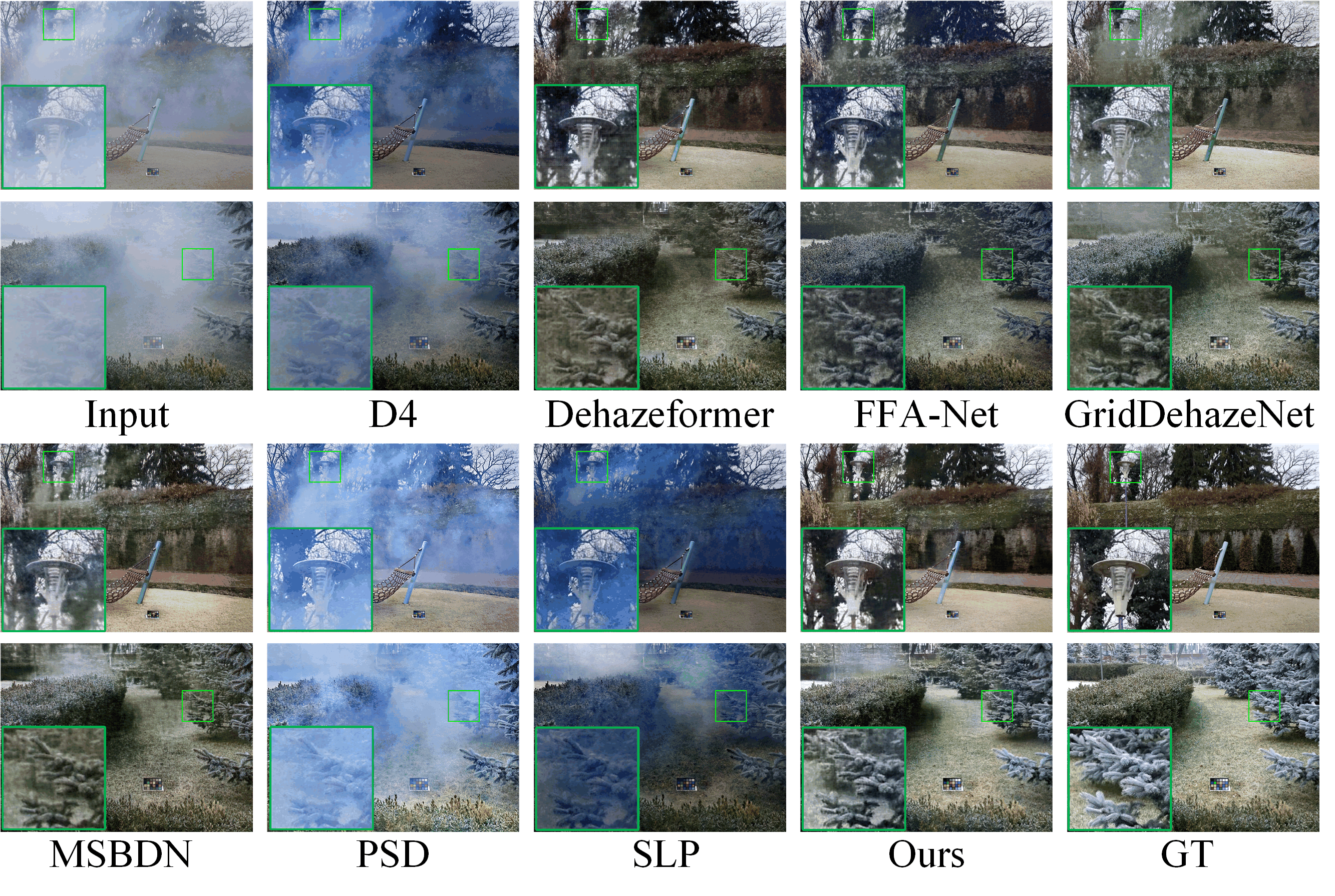}
\vspace{-2em}
\caption{Visual comparison of different dehazing methods on NH-HAZE dataset.(Zooming in can obtain a clearer view)}
\label{fig5}
\end{figure*}

\begin{figure*}
\centering
\includegraphics[width=1.0\linewidth]{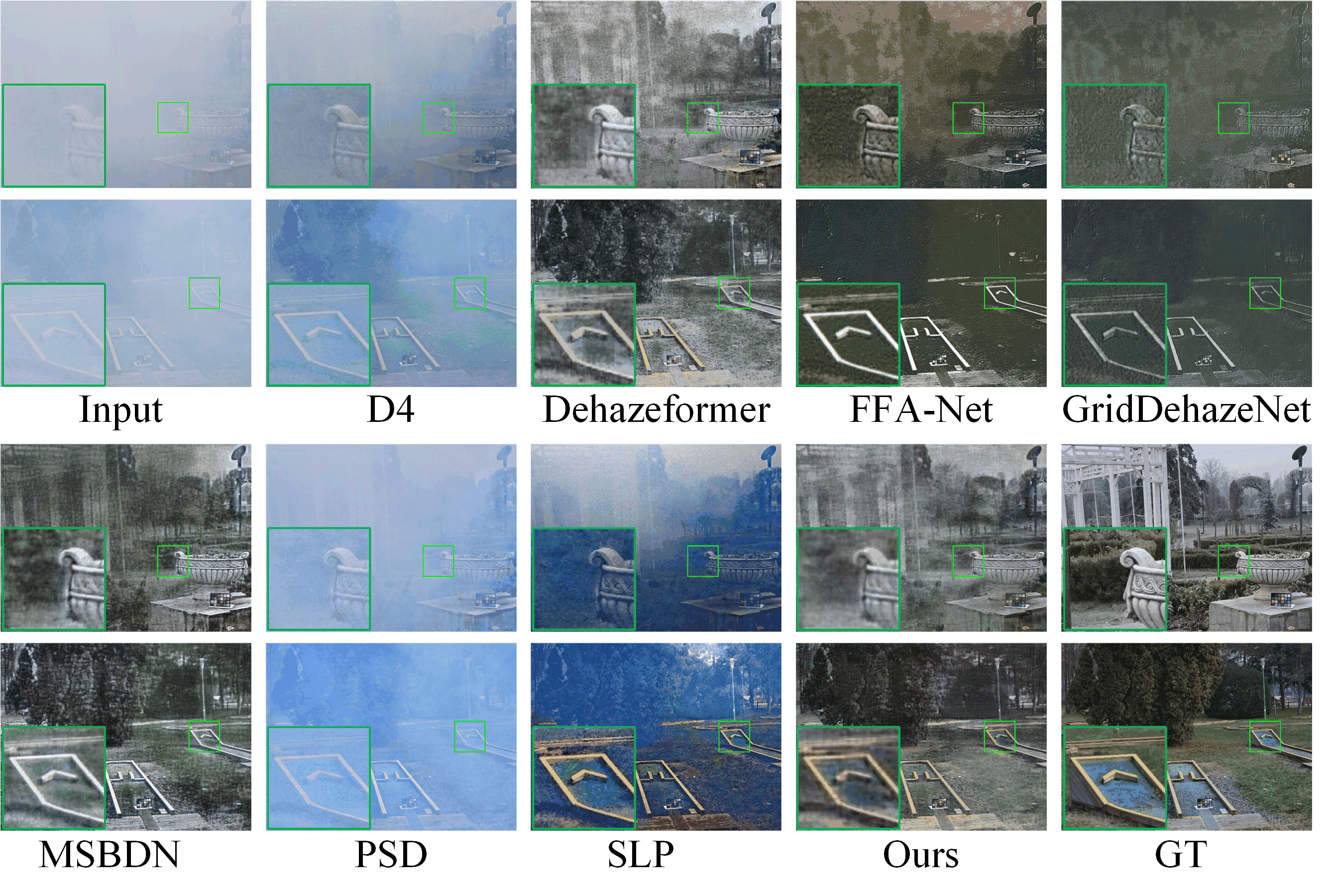}
\vspace{-2em}
\caption{Visual comparison of different dehazing methods on DENSE-HAZE dataset.(Zooming in can obtain a clearer view)}
\label{fig6}
\end{figure*}

\indent Table \ref{table1} presents a comparison of the quantitative measures obtained on the synthetic datasets. Table \ref{table1} shows that our method is second only to Dehazeformer in PSNR and SSIM in the RESIDE-6K test set, and also shows good performance in Entropy and LPIPS. It can be seen that the experimental results show that the proposed method has excellent dehazing performance on the synthetic dataset.\\
\textbf{\textit{Results on the  Real Datasets.}}
To further verify the dehazing ability of our method in real scenarios, we tested various methods on the uniform haze dataset DENSE-HAZE and the nonuniform haze NH-HAZE. The dehazing results are shown in Fig.\ref{fig5} and \ref{fig6}.

As shown in Fig.\ref{fig5}, the NH-HAZE dataset test shows that D4, PSD and SLP still have obvious haze, and methods such as GridDehazeNet, MSBDN and Dehazeformer have some problems of details blur and distortion.
As shown in Fig.\ref{fig6}, in the DENSE-HAZE dataset test, D4, PSD and SLP still have a lot of residual haze, while other methods have serious color deviation and relatively obvious noise. Overall, in terms of subjective evaluation, the comparison methods produce some problems such as blur, distortion or noise in both NH-HAZE and DENSE-HAZE scenes. In contrast, the images recovered by our method are closest to the real haze-free images in terms of both dehazing and color restoration, so our method has the best dehazing performance in subjective evaluation.

\indent Tables \ref{table2} and \ref{table3} present a comparison of the quantitative measures obtained on the real datasets NH-HAZE and DENSE-HAZE. As can be seen from Table \ref{table2}, in the DENSE-HAZE dataset, our method has the best values in PSNR and SSIM, and Entropy and LPIPS are only 0.133 and 0.0237 lower than Dehazeformer-T. Table \ref{table3} shows that in the NH-HAZE dataset, our method has the best values in PSNR, Entropy and LPIPS, and SSIM is only 0.0498 lower than Fourmer. The experimental results show that our method also has the best performance in objective evaluation.

\begin{table}
\scriptsize
\centering
\caption{Quantitative comparison of the proposed algorithm and different comparison methods on the DENSE-HAZE dataset.  \textbf{Bold} is the best, \textcolor{red}{Red} is the second.}
\label{table2}
\scalebox{1.15}{
\begin{tabular}{c||c||c|c|c|c}

\toprule
\multirow{3}{*}{\textbf{Methods}} & \multirow{3}{*}{\textbf{Venue\&  Year}} & \multicolumn{4}{c}{\textbf{DENSE-HAZE}} \\
\cmidrule(lr){3-6}
&& $PSNR\uparrow$ & \ $SSIM\uparrow$  & $Entropy\uparrow$ &  $LPIPS\downarrow$  \\
\midrule
GridDehazeNet \cite{c4} & \textit{ICCV'19} 
 & 14.49      & 0.4401         &5.8736            &0.7168  \\

FFA-Net \cite{c5} & \textit{AAAI'20}
& 15.17      & 0.3243    &  6.3659       &0.7069\\

MSBDN \cite{c6} & \textit{CVPR'20}
&  15.51      & 0.3478      &6.9032       &0.7197 \\

PSD \cite{PSD} & \textit{CVPR'21}
&9.73             &0.4345          &5.5030     &0.8184 \\

SG-Net \cite{SG-Net} & \textit{ACCV'22}
&14.91       &0.4641        & -            & -     \\

D4 \cite{D4} & \textit{CVPR'22}
&11.49       &0.4821      & 6.3123        &0.7555 \\

SLP \cite{SLP} & \textit{TIP'23}
&13.81       &0.4857         & 6.7357      &0.8489    \\

Dehazeformer-T \cite{T3} & \textit{TIP'23}
&15.52    &0.4635        &\textbf{7.093}         &\textbf{0.6459 }        \\

Fourmer\cite{fourmer} & \text{ICML 23}
&\textcolor{red}{15.95}      &\textcolor{red}{0.4917}         &-            &-            \\

\midrule
\rowcolor{gray!20} \textbf{Ours} & -
& \textbf{16.42  } & \textbf{0.5235} & \textcolor{red}{6.9600} & \textcolor{red}{0.66966}  
\\
\bottomrule
\end{tabular}}
\end{table}


\begin{table}
\scriptsize
\centering
\caption{Quantitative comparison of the proposed algorithm and different comparison methods on the NH-HAZE  dataset.  \textbf{Bold} is the best, \textcolor{red}{Red} is the second.}
\label{table3}
\scalebox{1.15}{
\begin{tabular}{c||c||c|c|c|c}

\toprule
\multirow{3}{*}{\textbf{Methods}} & \multirow{3}{*}{\textbf{Venue\&  Year}} & \multicolumn{4}{c}{\textbf{NH-HAZE}} \\
\cmidrule(lr){3-6}
&& $PSNR\uparrow$ & \ $SSIM\uparrow$  & $Entropy\uparrow$ &  $LPIPS\downarrow$  \\
\midrule
GridDehazeNet \cite{c4} &  \textit{ICCV’19}
& 17.23   & 0.5042  &7.2881     &0.3787 \\

FFA-Net \cite{c5} & \textit{AAAI’20}
& 18.09   & 0.5173  &7.2727      &\textcolor{red}{0.3635}\\

MSBDN \cite{c6} & \textit{CVPR’20 }
& 17.12   & 0.4539     & 7.3045       &0.3989 \\

PSD \cite{PSD} & \textit{CVPR'21}
&10.32     &0.5274     & 7.0658         &0.5247 \\

SG-Net \cite{SG-Net} & \textit{ACCV‘22}
&18.68   &0.6609        &-          &-     \\

D4 \cite{D4} & \textit{CVPR’22}
&12.66     &0.5072    & 7.1318         &0.5259 \\

SLP \cite{SLP} & \textit{TIP'23}
&15.84  &0.5956       &6.9696        &0.4687   \\

Dehazeformer-T \cite{T3} & \textit{TIP'23}
&18.73   &0.5326   &\textcolor{red}{7.3274}        &0.3649      \\

Fourmer\cite{fourmer} & \textit{ICML'23}
&\textcolor{red}{19.91}  &\textbf{0.7214}         &-            &-            \\

\midrule
\rowcolor{gray!20} \textbf{Ours} & -
& \textbf{20.10  } & \textcolor{red}{ 0.6716} & \textbf{ 7.5319} & \textbf{ 0.3210}  
\\
\bottomrule
\end{tabular}}
\end{table}

The subjective and objective experimental results described above demonstrate the superiority of our method and the effectiveness of the proposed interaction-guided global and local feature extraction models. We also attach in Table \ref{table5} the number of parameters for each deep learning model along with MACs.
\begin{table*}
\centering
\caption{Performance comparison of the comparative methods on the quantity of model parameters and MACs.}
\begin{adjustbox}{width=\textwidth}
\begin{tabular}{ c c c c c c c c c c c }
\toprule
 \multicolumn{1}{c}{\textbf{Methods}}  & \multicolumn{1}{c}{GridDehazeNet} & \multicolumn{1}{c}{FFA-Net}  & \multicolumn{1}{c}{MSBDN}   & \multicolumn{1}{c}{PSD}& \multicolumn{1}{c}{SG-Net}  & \multicolumn{1}{c}{D4}  & \multicolumn{1}{c}{SLP}  & \multicolumn{1}{c}{Dehazeformer-T}  & \multicolumn{1}{c}{Fourmer}  & \multicolumn{1}{c}{Our}\\
\midrule
\multirow{1}{*}{\textbf{Parameters}}        & 0.956M          & 4.456M  &31.35M   &6.21M     & 3.33M  &10.7M  &- & 0.686M &1.29M &22.87M\\ 
\multirow{1}{*}{\textbf{MACs}}           & 21.49G           & 287.8G   & 41.54G  & 143.91G     & 3.34G &2.25G    &-  &6.658G  &20.6G &17.52G\\ 
\midrule
\bottomrule
\end{tabular}
\label{table5}
\end{adjustbox}
\end{table*}

\subsection{Ablation Study}
In this section, we conducted ablation analysis on each component of the proposed method and verified the impact of each component on the performance of dehazing. Firstly, we build the basic network framework as the Base of the dehazing network, which consists of two branches of Transformer and CNN as Encoder then the features are summed up and pass through the Decoder composed of CNN. Then we add different modules to base, including:\\
(1) \textbf{Base+DownS}: downsample the image once before feeding it into the Transformer.\\
(2) \textbf{Base+DownS+FA}: downsample an input image before it is fed into the Transformer and outputs the features of the Transformer and CNN between each layer to be summed. \\
(3) \textbf{Base+FA+CPA}: between each layer, the Transformer and CNN feature outputs are summed and fed into the CPA to obtain a weight matrix, which is multiplied by the CNN outputs.\\
(4) \textbf{Ours}: Our model includes all the above blocks.

\begin{table}
\centering
\caption{Ablation studies with different modules on the NH-HAZE dataset.}
\label{table4}
\scalebox{1}{
\begin{tabular}{l||ccc||c|c}
\toprule
\textbf{Methods} & \textbf{DownS} & \textbf{FA} & \textbf{CPA} & PSNR & SSIM \\
\midrule
Base & - &  - &- &17.70 &0.5324\\
Base+DownS & \ding{52} & - &- & 19.14&0.5985  \\
Base+DownS+FA & \ding{52} &\ding{52} &-  &19.48 &0.6530 \\
Base+FA+CPA & - & \ding{52} & \ding{52} &18.96 &0.5608 \\
\midrule
\rowcolor{gray!20} \textbf{Ours} & \ding{52} & \ding{52} &  \ding{52}& \textbf{20.10}&\textbf{0.6716}  \\
\bottomrule
\end{tabular}
}
\end{table}
\indent For all models, we used L1 loss for image reconstruction and used the NH-HAZE dataset for training and testing in our ablation experiments. 
The quantitative evaluation results of the models described above are presented in Table \ref{table4}. All modules improved model performance compared to the Base model, demonstrating the overall effectiveness of our design. \\
(1) Base+DownS: Although Transformer and CNN tend to extract inconsistent feature information, there will still be many redundant features extracted by dual-branch feature extraction under the same dimension, which affects the performance of the model. Therefore, we add downsampling before the Transformer branch to increase the global receptive field of the Transformer branch and reduce the feature redundancy caused by the repeated extraction of double-branch features. 
Therefore, as shown in Table \ref{table4}, adding downsampling increased the PSNR and SSIM of the Base model from 17.70 and 0.5324 to 19.14 and 0.5985, respectively.\\
(2) Base+DownS+FA: On the basis of downsampling, the features of each layer of the two branches are extracted and added together, and then used to guide the CNN, so that the CNN branch has the ability to extract local features and global context, so that the model can better take into account global features and local features. Incorporating downsampling and feature addition is observed to increase the PSNR and SSIM from 19.14 and 0.5985 to 19.48 and 0.6530, respectively, compared to the model above.\\
(3) Base+CPA+FA: The model adds feature summation to the base model, and uses CPA to generate a weight matrix to guide the CNN, so that the model can focus on channels and pixels at the same time, which can better guide the CNN to capture information in the effective feature space. Without downsampling, these two modules improve the PSNR and SSIM of the Base model from 17.70 and 0.5324 to 18.96 and 0.5608.\\
(4) Ours: The full proposed model includes includes all the above blocks, and yielded the highest performance with PSNR and SSIM values of 20.10 and 0.6716, respectively, representing a PSNR increase of 2.40 dB compared with the Base model. These results clearly demonstrate the effectiveness of each module in the proposed model.
\begin{figure*}
\centering
\includegraphics[width=1.0\linewidth]{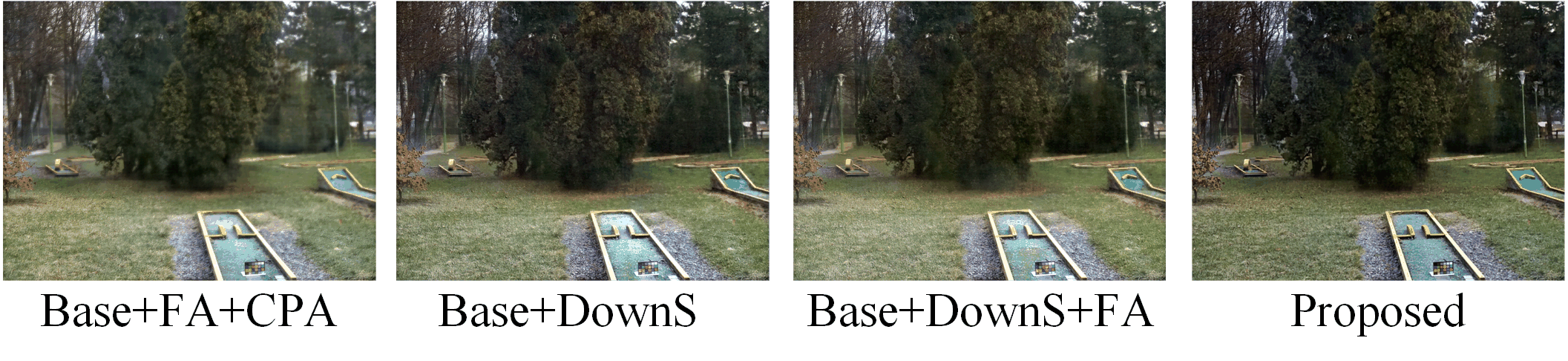}

\caption{Visual comparison of different dehazing methods on NH-HAZE dataset.(Zooming in can obtain a clearer view)}
\label{fig7}
\end{figure*}

The visual comparison of the ablation model is shown in Figure.\ref{fig7}. It can be seen from the figure that our complete model has better dehazing performance, the overall tone and brightness of the image are natural, the overall picture is full, and there is no edge blur or distortion. The visual comparison further verifies the effectiveness of the proposed module.

\section{Conclusion}
In this paper, we proposed an interaction-guided two-branch image dehazing network. The proposed model leverages the global and local feature extraction capabilities of a Transformer and CNN, respectively. It outputs the features between each layer for synthesis. The CPA block then considers these global and local features simultaneously to generate a weight matrix, which is multiplied by the outputs of the CNN to guide local feature extraction using global features. Additionally, the introduction of downsampling before the Transformer branch can effectively reduce computational complexity and increase the receptive field to improve overall model performance. The results of extensive experiments demonstrate that our method performs competitively in terms of both subjective and objective evaluations on synthetic and real datasets. Additionally, ablation analyses demonstrates the effectiveness of each module of the proposed method.\\

\noindent \textbf{Acknowledgment.} This work was supported by the Basic and Applied Basic Research of Guangdong Province under Grant 2023A1515140077, the Natural Science Foundation of Guangdong Province under Grant 2024A1515011880, the National Natural Science Foundation of China under Grant 62201149, the Guangdong Higher Education Innovation and Strengthening of Universities Project under Grant 2023KTSCX127.

\small

\bibliographystyle{splncs04}

\end{document}